\documentclass[10pt,twocolumn,letterpaper]{article}

\usepackage{iccv}      

\usepackage{multirow}
\usepackage{color}
\usepackage{colortbl}  
\usepackage{xcolor}
\definecolor{iccvblue}{rgb}{0.21,0.49,0.74}
\usepackage[pagebackref,breaklinks,colorlinks,allcolors=iccvblue]{hyperref}

\def\paperID{9032} 
\def\confName{ICCV}
\def\confYear{2025}

\title{
Reducing Unimodal Bias in Multi-Modal Semantic Segmentation with Multi-Scale Functional Entropy Regularization
}

\author{Xu Zheng$^{1,2}$\quad Yuanhuiyi Lyu$^{1}$\quad Lutao Jiang$^{1}$ \\ \quad Danda Pani Paudel$^{2}$ \quad Luc Van Gool$^{2}$ \quad Xuming Hu$^{1,3,}$\thanks{Corresponding author.} \\ \\
$^{1}$AI Thrust, HKUST(GZ) \quad $^{2}$INSAIT, Sofia University “St. Kliment Ohridski” \quad $^{3}$CSE, HKUST \\
}

\begin{document}
\maketitle
\begin{abstract}
Fusing and balancing multi-modal inputs from novel sensors for dense prediction tasks, particularly semantic segmentation, is critically important yet remains a significant challenge. One major limitation is the tendency of multi-modal frameworks to over-rely on easily learnable modalities, a phenomenon referred to as \textbf{unimodal dominance or bias}. This issue becomes especially problematic in real-world scenarios where the dominant modality may be unavailable, resulting in severe performance degradation. To this end, we apply a \textbf{simple but effective} plug-and-play regularization term based on functional entropy, which introduces no additional parameters or modules. This term is designed to intuitively balance the contribution of each visual modality to the segmentation results. Specifically, we leverage the log-Sobolev inequality to bound functional entropy using functional-Fisher-information. By maximizing the information contributed by each visual modality, our approach mitigates unimodal dominance and establishes a more balanced and robust segmentation framework. A multi-scale regularization module is proposed to apply our proposed plug-and-paly term on high-level features and also segmentation predictions for more balanced multi-modal learning. Extensive experiments on three datasets demonstrate that our proposed method achieves superior performance, \ie,\textbf{ +13.94\%, +3.25\% and +3.64\%}, without introducing any additional parameters.
\end{abstract}    
\begin{figure}[ht!]
    \centering
    \includegraphics[width=.95\linewidth]{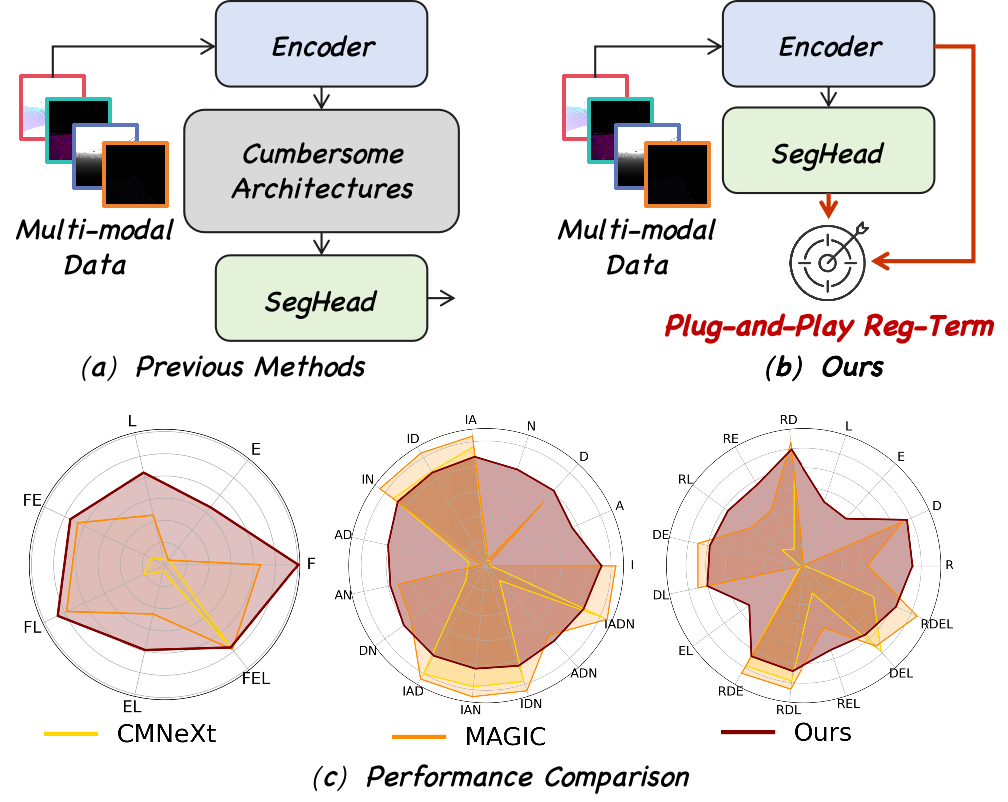}
    \caption{
    (a) Previous methods use complex architectures to fuse multi-modal data. (b) Our approach \textbf{\textit{simplifies}} the architecture and use plug-and-play regularization terms. 
    The results show significant improvements of ours than CMNeXt~\cite{zhang2023delivering} and MAGIC~\cite{zheng2025centering}.
    }
    \label{fig:cover}
\end{figure}

\section{Introduction}
\label{sec:intro}
Multi-modal dense prediction tasks, especially for semantic segmentation~\cite{xie2021segformer,chen2024frozen,zheng2023distilling,uncerawarecotrain,uncerteacher,EXACT,advercotrain,semanticdissty,ren2024sharing,ren2023masked,ren2024bringing}, rely on integrating complementary inputs from diverse sensors (\eg, RGB, depth, thermal, event cameras) to achieve robust scene understanding~\cite{zheng2023deep,zheng2025centering, zheng2024learning}. 
While such fusion promises resilience to real-world variability, a critical challenge persists: models often exhibit unimodal bias, disproportionately favoring one modality over others~\cite{zhangunderstanding,zheng2024anyseg}. 
For instance, in RGB-thermal segmentation, RGB inputs—common in training pipelines—often dominate predictions, even in low-light conditions where thermal data is crucial. Similarly, systems combining event cameras with conventional sensors may under-utilize the motion-aware edge information from event cameras in dynamic scenes, defaulting to static RGB inputs. This imbalance occurs when models prioritize the most easily learned or information-rich modality, neglecting others. As a result, performance degrades when the dominant modality is missing (\eg, RGB corruption) or unreliable (\eg, thermal sensor noise), posing risks in safety-critical applications like autonomous navigation.

Prior multi-modal methods often fail to explicitly regulate how modalities contribute to predictions~\cite{zhang2023delivering}, leaving models vulnerable to over-reliance on dominant inputs. Functional entropy addresses this gap by quantifying the uncertainty in a model’s dependence on modalities: \textit{a system with high entropy distributes its "attention" evenly across sensors, while low entropy indicates undue reliance on a single modality.} 
Inspired by functional entropy which has been successfully applied in visual question answering task~\cite{gat2020removing}, we apply this concept to the challenging multi-modal semantic segmentation.
However, directly applying the regularization is \textbf{\textit{non-trivial}} because the multi-modal semantic segmentation task presents significant challenges in enabling the model to learn pixel-wise multi-modal representations while balancing the contributions of multiple modalities, in contrast to the dual-modality and VQA task addressed in ~\cite{gat2020removing}.

%
In this paper, we apply the simple yet effective plug-and-play regularization terms at multiple scales of the segmentation model to balance various visual modalities while encourage equal multi-modal learning. 
As shown in Fig.~\ref{fig:cover} (c), framing modality balance as an entropy maximization problem, these regularization terms reduce bias and ensure equitable sensor utilization, crucial for segmentation accuracy when modalities degrade or fail.
Unlike complex architectures~\cite{zheng2025centering, zheng2024learning}, we draw from recent multi-modal learning advancements~\cite{wang2024qwen2}, which show that intricate connectors are increasingly unnecessary. This suggests that sophisticated fusion mechanisms are not required. We propose a simpler approach: instead of adding fusion modules, we treat all input modalities equally within a mini-batch, achieving fusion by averaging predictions, avoiding unnecessary complexity in the segmentation backbone.

To implement the proposed regularization term in multi-modal semantic segmentation tasks, we introduce a multi-scale regularization module that takes both multi-scale features and predictions into consideration. 
The prediction-level regularization is to ensure that predictions across different modalities are consistent, thereby encouraging the model to integrate information across modalities in a balanced manner. Meanwhile, the regularization with the multi-scale feature aims to enforce smoothness and balance, making sure that no single scale or modality disproportionately influences the final output.

Extensive experiments on real-world and synthetic benchmarks demonstrate the superior robustness and performance of our method compared to existing state-of-the-art approaches, achieving mIoU improvements of +13.94\%, +3.25\% and +3.64\% respectively. Additionally, we conduct series of ablation studies to show how the proposed regularization terms work.
Our contributions are as follows:
\textbf{(I)} We introduce functional entropy-based regularization to mitigate unimodal bias and ensure balanced multi-modal learning, reducing reliance on any single modality.  
\textbf{(II)} We propose a straightforward multi-modal fusion framework that averages input modalities within a mini-batch, avoiding complex fusion modules and simplifying the segmentation backbone.  
\textbf{(III)} Our regularization terms are applied at both feature and prediction levels, effectively balancing multi-modal training. Extensive experiments show robust performance across various benchmarks.

\section{Related Work}

\subsection{Multi-modal Semantic Segmentation}
Multi-modal Semantic Segmentation seeks to combine RGB with complementary modalities such as depth~\cite{lyu2024omnibind, lyu2024unibind, lyu2024image, wang2020learning, zhou2020rgb, wang2020deep, cao2021shapeconv, chen2021spatial, ying2022uctnet, lee2022spsn, cong2022cir, ji2022dmra, wang2022learning, song2022improving}, thermal~\cite{shivakumar2020pst900, zhang2021abmdrnet, wu2022complementarity, liao2022cross, zhou2023mmsmcnet, xie2023cross, chen2022modality, pang2023caver, hui2023bridging, zhang2023efficient,SegmentRBGT}, events~\cite{alonso2019ev, zhang2021issafe, zheng2024eventdance, cao2023chasing, zhou2024exact}, and LiDAR~\cite{zhuang2021perception, yan20222dpass, wang2022multimodal, li2022deepfusion, borse2023x, zhang2023mx2m, liu2022camliflow, li2023mseg3d}. Advances in sensor technology have driven significant progress in multi-modal fusion~\cite{,cao2023chasing,lyu2024omnibind,goodsam,goodsam++,magic++,memorysam,MMSSBench,OmniSAM,EventDance++}, evolving from dual-modality to comprehensive multi-modal systems, like MCubeSNet~\cite{liang2022multimodal}, which improves scene understanding by utilizing richer sensor data.

From an architectural standpoint, multi-modal fusion models are generally classified into three categories: separate branches~\cite{broedermann2022hrfuser, wei2023mmanet, zhang2021abmdrnet, man2023bev}, joint branches~\cite{wang2022multimodal, chen2021spatial}, and asymmetric branches~\cite{zhang2023delivering, zhang2022cmx}. A common approach in these models is to treat RGB as the primary modality, while auxiliary sensors provide additional information. For instance, CMNeXT~\cite{zhang2023delivering} prioritizes RGB, incorporating other sensors to supplement the data. However, RGB alone may be insufficient, particularly in challenging conditions like low light or nighttime. This limitation highlights the need for more robust fusion models that leverage the strengths of multiple modalities, minimizing dependence on any single sensor. In this context, Liu \etal~\cite{liu2024fourier} introduced the concept of modality-incomplete scene segmentation, addressing deficiencies at both the system and sensor levels. 



\begin{figure*}[h!]
    \centering
    \includegraphics[width=\textwidth]{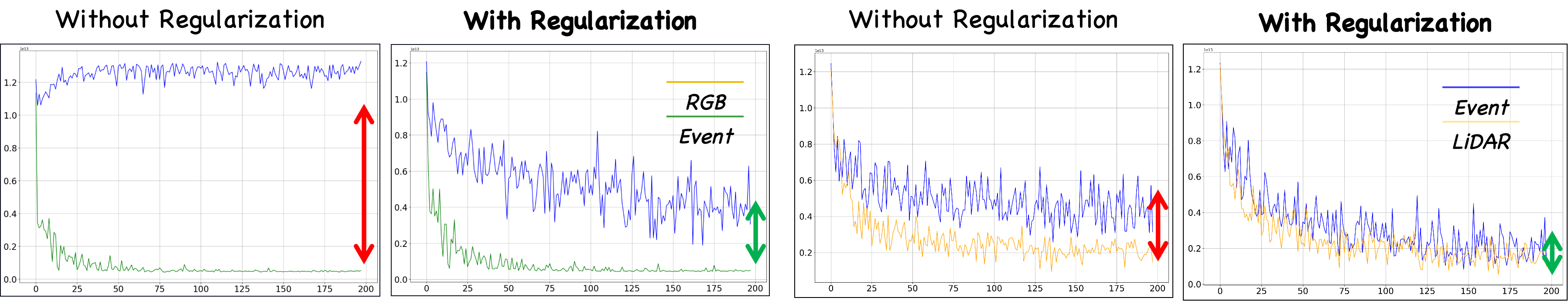}
    \caption{Visualization of Fisher information during training with real-world MUSES (right) and synthetic DELIVER (left) datasets. Adding a regularization term \textbf{\textit{reduces the gap}} between modality Fisher information values. With regularization, the model utilizes both input modalities, resulting in \textbf{\textit{closer modality distances.}} Without regularization, the model relies more on RGB in the first two rows and Event in the last two, increasing the modality distance (red arrows). 
    }
    \label{fig:fisher_info}
\end{figure*}

\subsection{Unimodal Bias}
As shown in Fig.~\ref{fig:fisher_info}, a key challenge in multi-modal tasks, especially for multi-modal semantic segmentation, is unimodal bias, where where models favor one modality over others, leading to suboptimal performance. 
For instance, as shown in Tab.~\ref{Tab:MUSES}, the frame camera has a dominant effect on the performance of most methods. Specifically, when the frame camera is included in the evaluation, the performance significantly increases. However, when it is missing, the performance drops sharply.
This bias arises when models prioritize the most easily learnable or information-rich modality under specific conditions, neglecting the complementary benefits of other modalities. 
This can lead to a significant degradation in performance when the dominant modality is missing or unreliable~\cite{zheng2024learning, zheng2025centering, liu2024fourier}, such as in cases of RGB corruption or thermal sensor noise, which can be particularly problematic for safety-critical applications like autonomous navigation.
Several strategies have been proposed to address unimodal bias in multi-modal learning~\cite{zheng2024learning, zheng2025centering}. Traditional multi-modal fusion models often do not explicitly regulate how each modality contributes to the final predictions, which can lead to over-reliance on a single input. Recent works, such as MAGIC~\cite{zheng2025centering} and Any2Seg~\cite{zheng2024learning} have sought to address this limitation by introducing well-designed training objectives. 


In multi-modal learning, methods~\cite{zhangunderstanding, zheng2024anyseg} often draw from information theory, particularly the concept of functional entropy~\cite{gat2020removing}. Entropy-based approaches quantify a model's uncertainty in its reliance on different modalities. A system with high entropy distributes its "attention" evenly across modalities, while low entropy signals an over-reliance on a single modality.
Building on the application of functional Fisher information in visual question answering tasks~\cite{gat2020removing}, directly applying this approach to semantic segmentation presents challenges. Unlike visual question answering, which focuses on interpreting questions based on a single modality, segmentation tasks require the integration of both spatial and contextual information across multiple modalities.
To address this, we introduce multi-scale regularization terms at both the feature and prediction levels. These regularization terms provide a principled framework to mitigate unimodal bias, encouraging a more balanced utilization of all available sensors. This ensures that models maintain accuracy even in scenarios where certain modalities degrade or fail—an essential consideration in environments with unreliable or variable sensor performance.

\begin{figure*}[ht!]
    \centering
    \includegraphics[width=\textwidth]{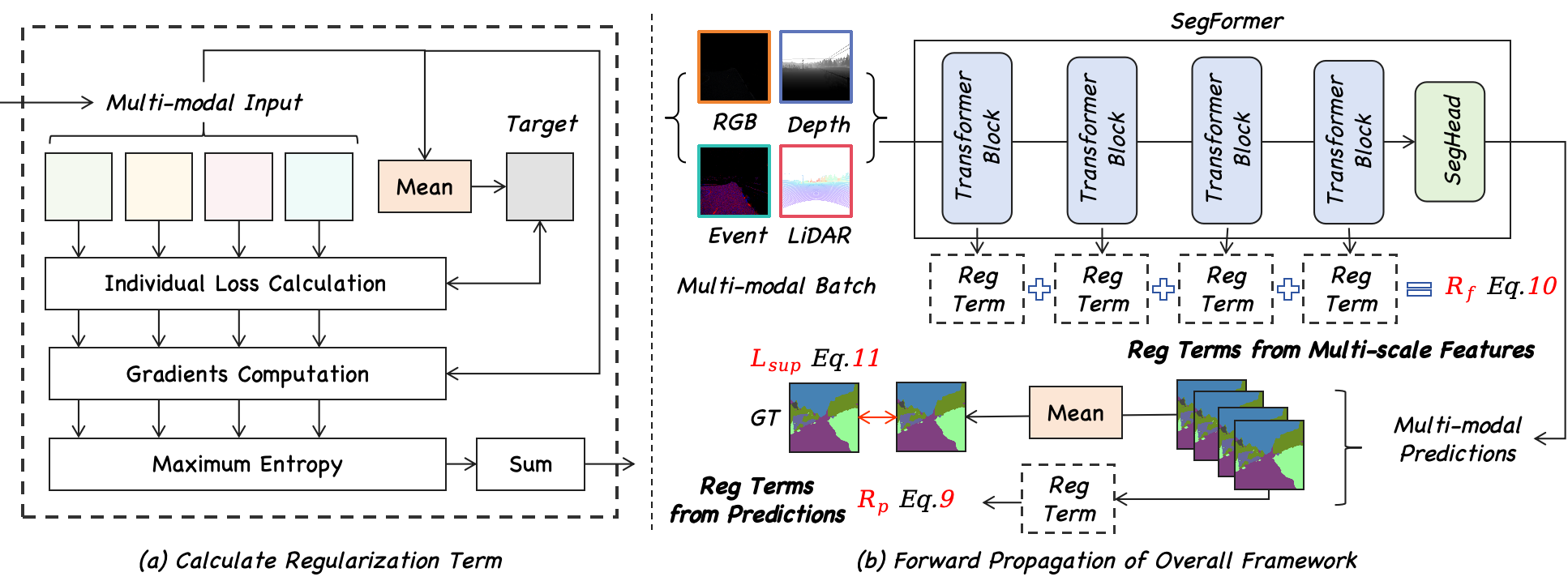}
    \caption{(a) The proposed regularization term for multi-modal input and (2) overall framework of our proposed multi-scale regularization terms' implementation. 
    }
    \label{fig:overall}
\end{figure*}

\section{Method}
\label{sec:method}

This paper focuses on multi-modal visual data from novel sensors, where each instance in a data batch \(x\) consists of multiple modalities. For instance, in multi-modal semantic segmentation~\cite{zhang2023delivering}, RGB images provide color and texture information, depth maps capture geometric details, event cameras offer motion cues, and LiDAR provides accurate point clouds. A mini-batch of data from these modalities can be represented as:
$x = \{ x_{\text{r}}, x_{\text{d}}, x_{\text{e}}, x_{\text{l}} \}$,
where each modality resides in its own Euclidean space, i.e., \(x_i \in \mathbb{R}^{d}\). Here, \(x_{\text{r}}\) represents the RGB, \(x_{\text{d}}\) represents the depth , \(x_{\text{e}}\) represents the event, and \(x_{\text{l}}\) represents the LiDAR modality, with each \(x_i\) being a vector in a \(d\)-dimensional space.
The goal of multi-modal semantic segmentation is to optimize the parameters \( w \) of a function that assigns a probability score to each possible label \( y \) given the input mini-batch \( x \). 

\subsection{Reducing Unimodal Dominance / Bias}

In the context of multi-modal semantic segmentation, existing methods~\cite{zheng2025centering, zheng2024learning, zhang2023delivering} introduce additional modules and parameters to encourage balanced training across multiple visual modalities. However, neural networks often tend to overly rely on a single modality, leading to significant challenges in real-world scenarios.
For instance, as shown in Tab.~\ref{Tab:MUSES}, the frame camera has a dominant effect on the performance of most methods. Specifically, when the frame camera is included in the evaluation, the performance significantly improves. However, its absence results in a sharp decline in performance. This phenomenon, commonly referred to as \textbf{\textit{unimodal dominance}} or \textit{bias}, occurs when the learned function disproportionately emphasizes the most easily learnable modality, potentially neglecting critical information from other modalities.

\subsubsection{Functional Entropy}
\label{Functional Entropy}
In multi-modal learning, functional entropy quantifies the uncertainty and variability in the model's multi-modal predictions and representation~\cite{gat2020removing}. 
Let \( f(x) \) represent the model’s prediction for a multi-modal input mini-batch \( x = \{ x_{\text{r}}, x_{\text{d}}, x_{\text{e}}, x_{\text{l}} \} \), where \( x \in \mathbb{R}^d \) includes RGB, Depth, LiDAR, and Event data. The functional entropy of a non-negative function \( f(x) \geq 0 \) is defined as:
\begin{align}
\setlength{\abovedisplayskip}{3pt}
\setlength{\belowdisplayskip}{3pt}
  & \text{Ent}_\mu(f) = \int_{\mathbb{R}^d} f(x) \log f(x) \, d\mu(x) - \notag \\ 
  & \quad \left( \int_{\mathbb{R}^d} f(x) \, d\mu(x) \right) \log \left( \int_{\mathbb{R}^d} f(x) \, d\mu(x) \right),
\end{align}
where \( f(x) \) represents the model's predictions in the form of segmentation logits and multi-scale features, which reflect how the model combines different modalities. Here, \( \mu \) represents a probability measure defined over the Euclidean space \( \mathbb{R}^d \). This measure describes the distribution of the multi-modal inputs, quantifying how likely each point (pixel in prediction logits or element from high-level features) in the input space is, typically based on a distribution over \( \mathbb{R}^d \).
As shown in Fig.~\ref{fig:fisher_info}, a low entropy suggests over-reliance on a single modality (\eg, RGB), while higher entropy indicates better modality integration (\eg, RGB, Depth, LiDAR, Event data).

During training, we apply random modality dropout to expose the model to diverse input combinations.
Let \( x' = \{ x_{1}, x_{2}, \dots, x_{i} \} \) represent the available modalities during training, where \( 1, 2, \dots, i \in \{r, d, e, l\} \) are a subset of modalities (RGB, Depth, Event, LiDAR), and \( n \leq 4 \). This accounts for missing modalities by excluding corresponding terms from the input.
For a non-negative function \( f(x') \), the functional entropy is bounded as:
\begin{equation}
\setlength{\abovedisplayskip}{3pt}
\setlength{\belowdisplayskip}{3pt}
    \text{Ent}_\mu(f) \leq \frac{1}{2} \int_{\mathbb{R}^d} \frac{\|\nabla f(x')\|^2}{f(x')} \, d\mu(x'),
\end{equation}
where \( \|\nabla f(x')\| \) is the gradient of \( f(x') \). Large Fisher information values indicate a model’s dependence on specific modalities (\eg, RGB), suggesting a need for better modality integration, even with missing inputs.

Different from ~\cite{gat2020removing}, which focuses solely on the image-level visual question answering task using only image and text modalities, we aim to address the \textbf{\textit{more complex pixel-level understanding task by incorporating multiple visual modalities}}. Simply increasing the utilization of the visual modality in visual question answering is straightforward; however, when multiple visual modalities are integrated with pixel-level tasks such as semantic segmentation, the problem becomes far more challenging. These challenges stem not only from improving the utilization of the visual modality but also from effectively balancing multiple modalities.
To better handle multi-modal data, we decompose the overall functional entropy across modalities, allowing the model to independently assess the contribution of each modality. This is crucial when some modalities are missing during training or inference. By isolating the impact of each modality, the model can adapt to modality-missing scenarios more effectively.
For a multi-modal segmentation model, the functional entropy is bounded as:
\begin{equation}
\setlength{\abovedisplayskip}{3pt}
\setlength{\belowdisplayskip}{3pt}
\text{Ent}_\mu(f^x) \leq \sum_{i=1}^n \text{Ent}_{\mu_i}(f(x_i)),
\end{equation}
where \( \mu_i \) is the marginal distribution for modality \( x_i \). This decomposition enables the model to focus on each modality’s role, improving efficiency and flexibility. As a result, the model provides more accurate predictions even when some modalities are missing or incomplete.


\subsubsection{Regularization}
To enhance multi-modal integration and prevent modality dominance, the regularization term is:
\begin{align}
\setlength{\abovedisplayskip}{3pt}
\setlength{\belowdisplayskip}{3pt}
\mathcal{R} = \lambda \sum_{i=1}^n \left( \int_{\mathbb{R}^{d_i}} \frac{\|\nabla_{x_i} \text{CE}(p_w(\cdot | x_i), p_w(\cdot | x))\|^2}{\text{CE}(p_w(\cdot | x_i), p_w(\cdot | x))} \, d\mu_i(x_i) \right)^{-1}
\label{eq6}
\end{align}
where \( \lambda \) controls regularization strength.
This regularization term serves to promote smoother predictions and ensures a more balanced contribution from all modalities. By discouraging over-reliance on any single modality—such as the frame camera, which may underperform in low-light conditions—it encourages the use of complementary modalities. For instance, depth data aids in capturing geometric information, while LiDAR offers enhanced precision.
However, directly estimating functional entropy from the equation presents practical challenges. To address this, we leverage the log-Sobolev inequality for Gaussian measures, which provides a bound on functional entropy via functional Fisher information. This approach implicitly encourages the maximization of information across modalities. To implement this, we utilize the inverse of the information derived from Equation~\ref{eq6}, thereby promoting a more equitable contribution from each modality. The efficacy of this regularization method is demonstrated in Fig.~\ref{fig:fisher_info}, where it improves the stability and robustness of the multi-modal segmentation process.

\subsection{Overall Framework}
\label{Overall framework}

\subsubsection{Overview}
The concepts of functional entropy and functional Fisher information provide a theoretical foundation for analyzing and improving multi-modal semantic segmentation. By quantifying how different modalities are integrated and balanced, these measures address modality dominance, leading to more robust and generalized predictions. This approach is particularly important in multi-modal tasks, where leveraging the complementary strengths of each modality is essential for accurate segmentation.

Recent advancements in multi-modal learning models, such as Qwen2-VL~\cite{wang2024qwen2}, have shown that the need for complex connectors in multi-modal architectures is diminishing. Early models, such as ViLT, demonstrated that simple linear projections could effectively process image data for Transformers. More recent lightweight models have adopted multi-layer perceptrons (MLPs) to handle visual inputs, sometimes eliminating dedicated visual encoders altogether. These trends suggest that sophisticated fusion mechanisms are not necessarily required for effective multi-modal integration.
Building on these insights, we propose a straightforward approach: instead of using additional fusion modules, we treat all input visual modalities equally within a single mini-batch. Multi-modal fusion is achieved simply by taking the mean of the input modalities, \textit{\textbf{without complicating the backbone semantic segmentation framework}.}

In the forward pass, the multi-modal mini-batch \( x = \{ x_{\text{r}}, x_{\text{d}}, x_{\text{e}}, x_{\text{l}} \} \) is processed directly by the encoder within the backbone, producing multi-scale, high-level feature representations \( \{f_r^j, f_d^j, f_l^j, f_e^j\}_{j=1}^4 \) as well as the prediction segmentation maps \( \{p_r, p_d\} \) for the respective modalities. This operation is formulated as:
\begin{equation}
\setlength{\abovedisplayskip}{3pt}
\setlength{\belowdisplayskip}{3pt}
\{f_r^j, f_d^j\}_{j=1}^4, \{p_r, p_d\} = F(\{r, d\}),
\end{equation}
where \( j \) corresponds to the feature level derived from the \( j \)-th transformer block of the encoder. The final prediction \( p \) for the input multi-modal mini-batch \( x \) is obtained by averaging the individual modality predictions: $p = \text{Mean}(p_r, p_d).$


\begin{table}[t!]
\caption{Results with dual (RGB-Depth) modalities on DELIVER dataset. $^{\dagger}$ denotes the MMS method from ~\cite{liu2024fourier} and $^{\ddag}$ follows ~\cite{lee2023multimodal}.}
\label{tab:delvier_RD}
\resizebox{\linewidth}{!}{
\begin{tabular}{l|c|c|c|c}
\midrule
\multirow{2}{*}{Method} & \multicolumn{3}{c|}{RGB-Depth Fusion}                             & \multirow{2}{*}{Mean} \\ \cmidrule{2-4}
         & \multicolumn{1}{c|}{RGB\textcolor{gray!40}{-Depth}}    & \multicolumn{1}{c|}{\textcolor{gray!40}{RGB-}Dpeth}  & RGB-Depth  &                       \\ \midrule
CMNeXt~\cite{zhang2023delivering}                   & 1.60   & \multicolumn{1}{c|}{1.44}   & 63.58 & 22.81                 \\ 
CMNeXt$^{\ddag}$ & 32.97 & 48.53 & 61.93 & 47.81 \\
CMNeXt$^{\dagger}$ & \underline{53.39} & 53.73 & 62.24 & \underline{56.45} \\
MultiMAE~\cite{bachmann2022multimae} &24.60& 38.55 & 58.94 & 40.70\\
MultiMAE$^{\ddag}$ & 19.24 & 42.54 & 56.62 & 39.47 \\ 
MultiMAE$^{\dagger}$ & 52.46 & 45.62 & 58.68 & 52.25\\
FPT~\cite{liu2024fourier} & 50.73 & 39.60 & 57.38 & 49.24 \\
MAGIC~\cite{zheng2025centering}  & 37.26  & \underline{59.02}  & \textbf{66.89} & 54.39                 \\ \midrule
\rowcolor{gray!10} Ours    & \textbf{55.04}  & \textbf{59.60}  & \underline{66.46} & \textbf{60.37} \\ 
\bottomrule
\end{tabular}}
\end{table}

\begin{table*}[ht!]
\renewcommand{\tabcolsep}{14pt}
\caption{Results of segmentation validation with three modalities (frame: F, event: E, and LiDAR: L) on MUSES~\cite{brodermann2025muses}.}
\renewcommand{\tabcolsep}{20pt}
\resizebox{\linewidth}{!}{
\begin{tabular}{c|ccccccc|c}
\midrule
\multirow{2}{*}{Method} & \multicolumn{7}{c}{Training with All Modalities and Evaluation with Different Combinations} & \multirow{2}{*}{Mean}  \\ \cmidrule{2-8}
& F & E & L & FE & FL & EL & FEL &  \\ \midrule
CMX~\cite{zhang2023cmx} & 2.52 & 2.35 & 3.01 & 41.15 & 41.25 & 2.56 & 42.27 & 19.30 \\ \cmidrule{1-2} \cmidrule{3-9} 
CMNeXt~\cite{zhang2023delivering} & 3.50 & 2.77 & 2.64 & 6.63 & 10.28 & 3.14 & 46.66 & 10.80 \\ \cmidrule{1-2} \cmidrule{3-9} 
MAGIC~\cite{zheng2025centering} & 43.22 & 2.68 & 22.95 & 43.51 & 49.05 & 22.98 & 49.02 & 33.34 \\ \cmidrule{1-2} \cmidrule{3-9} 
Any2Seg~\cite{zheng2024learning} & 44.40 & 3.17 & 22.33 & 44.51 & 49.96 & 22.63 & \textbf{50.00} & 33.86 \\ \cmidrule{1-2} \cmidrule{3-9} 
\rowcolor{gray!10} Ours & \textbf{60.33} & \textbf{33.15} & \textbf{42.59} & \textbf{47.19} & \textbf{53.47} & \textbf{39.59} & 47.89 & \textbf{47.80}  \\ 
\bottomrule
\end{tabular}}
\label{Tab:MUSES}
\end{table*}

\begin{table*}[ht!]
\renewcommand{\tabcolsep}{4pt}
\caption{Results of segmentation validation with four modalities (RGB: R, Depth: D, Event: E, and LiDAR: L) on DELIVER~\cite{zhang2023delivering}.}
\resizebox{\linewidth}{!}{
\begin{tabular}{c|ccccccccccccccc|c}
\midrule
\multirow{2}{*}{Method} & \multicolumn{15}{c}{Training with All Modalities and Evaluation with Different Combinations} & \multirow{2}{*}{Mean}\\ \cmidrule{2-16}
 & R & D & E & L & RD & RE & RL & DE & DL & EL & RDE & RDL & REL & DEL & RDEL & \\ \midrule
CMNeXt~\cite{zhang2023delivering} & 0.86 & 0.49 & 0.66 & 0.37 & 47.06 & 9.97 & 13.75 & 2.63 & 1.73 & 2.85 & 59.03 & 59.18 & 14.73 & 59.18 & 39.07 & 20.77 \\ \midrule
MAGIC~\cite{zheng2025centering} & 32.60 & 55.06 & 0.52 & 0.39 & 63.32 & 33.02 & 33.12 & 55.16 & 55.17 & 0.26 & 63.37 & 63.36 & 33.32 & 55.26 & 63.40 & 40.49 \\ \midrule
Any2Seg~\cite{zheng2024learning} & 39.02 & 60.11 & 2.07 & 0.31 & 68.21 & 39.11 & 39.04 & 60.92 & 60.15 & 1.99 & 68.24 & 68.22 & 39.06 & 60.95 & 68.25 & 45.04 \\ \midrule
\rowcolor{gray!10} Ours & \textbf{55.56} & \textbf{57.74} & \textbf{32.40} & \textbf{34.26} & 59.69 & \textbf{47.70} & \textbf{47.78} & 48.99 & 50.26 & \textbf{34.37} & 53.42 & 54.12 & \textbf{45.24} & 47.33 & 51.65 & \textbf{48.29} \\
\bottomrule
\end{tabular}}
\label{Tab:DELIVER}
\end{table*}

\subsubsection{Multi-scale Regularization Module}
To implement the proposed regularization term in multi-modal semantic segmentation tasks, we introduce a multi-scale regularization module that takes both multi-scale features and predictions into consideration. This module encourages balanced learning at both the feature and prediction levels, ensuring that all modalities contribute equally to the final segmentation output across all scales.

\noindent \textbf{Regularization with Predictions:}
We introduce a regularization term at the prediction level. The objective is to ensure that predictions at different scales are consistent, thereby encouraging the model to integrate information across modalities in a balanced manner. The regularization term for predictions can be written as:
\begin{align}
\setlength{\abovedisplayskip}{3pt}
\setlength{\belowdisplayskip}{3pt}
\mathcal{R}_{\text{p}} = \lambda_{\text{p}} \sum_{i=1}^n \left( \int_{\mathbb{R}^{d_i}} \frac{\|\nabla_{x_i} \text{CE}(p, gt)\|^2}{\text{CE}(p, gt)} \, d\mu_i(x_i) \right)^{-1},
\end{align}
where \( gt \) represents the ground truth for the semantic segmentation, and \( p \) is the predicted output. \( \lambda_{\text{p}} \) is a regularization parameter controlling the strength of the prediction-level regularization. This term penalizes inconsistencies in predictions across different modalities, encouraging the model to produce balanced outputs.

\noindent \textbf{Regularization with Features:}
As Fig.~\ref{fig:overall} shows, we take the multi-scale features \( \{f_r^j, f_d^j\}_{j=1}^4 \) from the segmentation backbone model to perform feature-level functional entropy-based regularization. This regularization enforces smoothness and balance across the multi-scale features, ensuring that no single scale or modality disproportionately influences the final output. The regularization term for features can be formulated as:
\begin{align}
\mathcal{R}_{\text{f}} = \lambda_{\text{f}} \sum_{j=1}^4 \sum_{i=1}^n \left( \int_{\mathbb{R}^{d_i}} \frac{\|\nabla_{f_i} \text{CE}(\{f_r^j, f_d^j\}, f_{m}^j)\|^2}{\text{CE}(\{f_r^j, f_d^j\}, f_{m}^j)} \, d\mu_i(f_i^j) \right)^{-1},
\end{align}
where \( \lambda_{\text{f}} \) is a regularization parameter that controls the strength of the feature-level regularization, and \( f_j \) represents the features extracted at the \( j \)-th scale. The regularization encourages consistency and smoothness in feature representations across different scales, preventing any one scale from dominating the learning process.

\subsection{Training Objectives}

The supervised loss uses the cross-entropy between the predicted labels \( p \) and the ground truth \( gt \), as follows:
\begin{equation}
\setlength{\abovedisplayskip}{3pt}
\setlength{\belowdisplayskip}{3pt}
    \mathcal{L}_{\textit{sup}} = \text{CE}(p, gt).
\end{equation}
To ensure that the model generalizes effectively and integrates multi-modal information in a balanced manner, the overall training objective is formulated as a combination of the supervised loss and the regularization terms:
\begin{equation}
\setlength{\abovedisplayskip}{3pt}
\setlength{\belowdisplayskip}{3pt}
    \mathcal{L} = \mathcal{L}_{\textit{sup}} + \mathcal{R}_{\text{p}} + \mathcal{R}_{\text{f}},
\end{equation}
where \( \mathcal{R}_{p} \) and \( \mathcal{R}_{f} \) are the regularization terms that ensure consistency and smoothness in the model’s predictions and features, respectively.

\begin{figure}[t!]
    \centering
    \includegraphics[width=\linewidth]{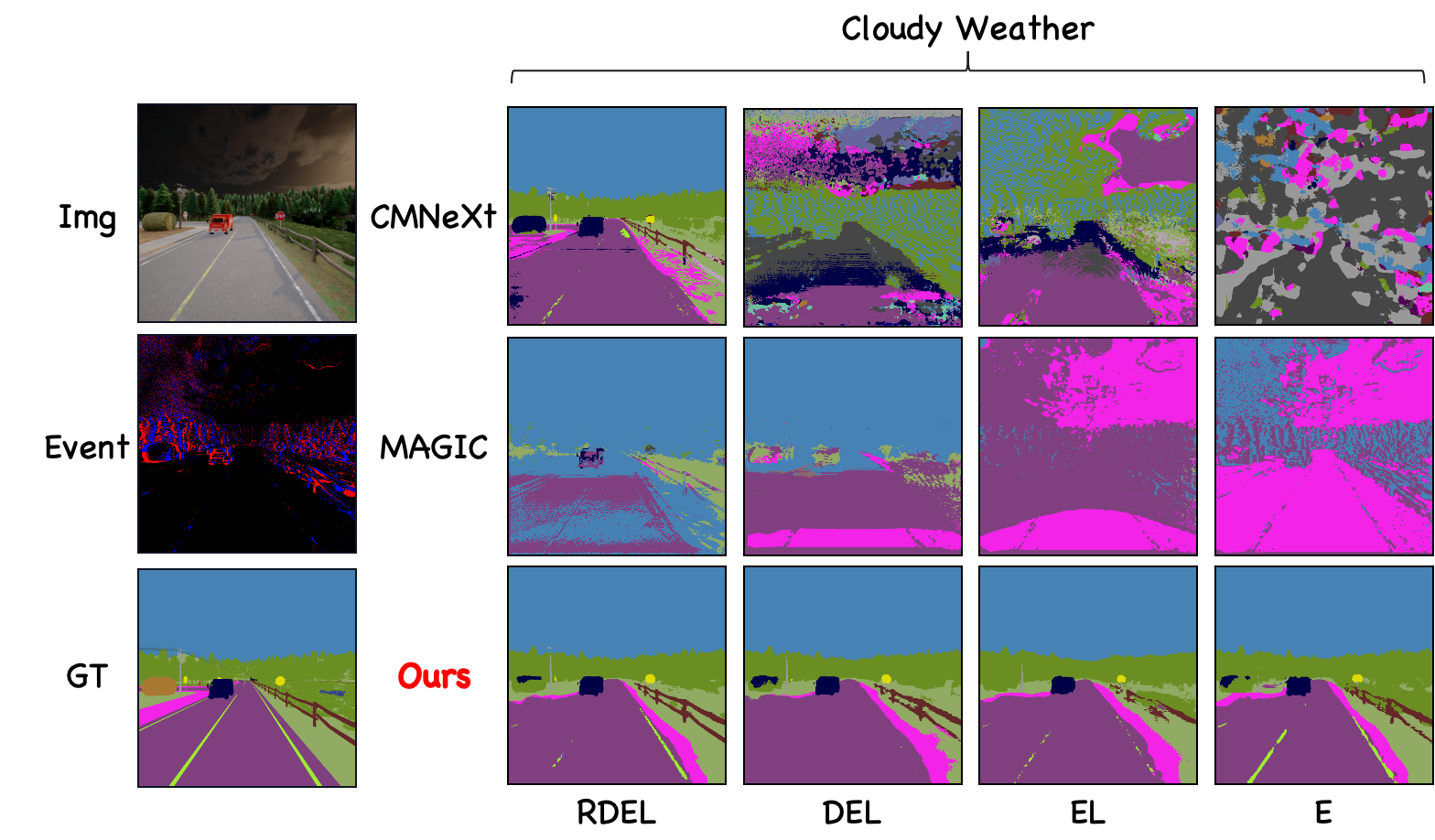}
    \caption{Visualization on DELIVER~\cite{zhang2023delivering} with~\cite{zhang2023delivering} and~\cite{zheng2025centering}.}
    \label{fig:visualization}
\end{figure}

\section{Experiments}
\noindent \textbf{Datasets.} 
We evaluate our method using synthetic and real-world multi-sensor datasets. The MUSES dataset~\cite{brodermann2024muses} provides data from a high-resolution frame camera, event camera, and MEMS LiDAR, supporting robust multimodal semantic segmentation with high-quality 2D panoptic labels for benchmarking. The DELIVER dataset~\cite{zhang2023delivering} includes RGB, depth, LiDAR, and event data across 25 categories, capturing diverse environmental conditions and sensor failures. MCubeS~\cite{liang2022multimodal} is a material segmentation dataset with 20 categories and pairs of RGB, Near-Infrared (NIR), DoLP, and AoLP images.
\noindent \textbf{Implementation Details.} All experiments were conducted on 8 NVIDIA GPUs. The initial learning rate was set to \(6 \times 10^{-5}\) and adjusted using a polynomial decay strategy with a power of 0.9 over 200 epochs. More detals can be found in the \textit{suppl. mat.}.

\begin{table*}[ht!]
\caption{Results of segmentation validation with four modalities (I: Image, Aolp: A, Dolp: D, and Nir: N) on MCubeS~\cite{liang2022multimodal}.}
\label{Tab:MCubeS}
\renewcommand{\tabcolsep}{6pt}
\resizebox{\textwidth}{!}{
\begin{tabular}{c|ccccccccccccccc|c}
\toprule
\multirow{2}{*}{Method} & \multicolumn{15}{c}{Training with All Modalities and Evaluation with Different Combinations} & \multirow{2}{*}{Mean}\\ \cmidrule{2-16}
 & I & A & D & N & IA & ID & IN & AD & AN & DN & IAD & IAN & IDN & ADN & IADN & \\ \midrule
CMNeXt~\cite{zhang2023delivering} & 1.86 & 1.54 & 2.51 & 2.28 & 47.96 & 43.67 & 45.90 & 6.99 & 7.58 & 9.95 & 50.07 & 48.77 & 48.83 & 8.06 & 51.54 & 25.17 \\ \midrule
MAGIC~\cite{zheng2025centering} & 51.91 & 0.32 & 34.52 & 2.66 & 52.24 & 52.16 & 52.57 & 1.98 & 36.01 & 37.09 & 52.48 & 52.82 & 52.73 & 37.57 & 53.01 & 38.00 \\ \midrule
\rowcolor{gray!10} Ours & 46.21 & \textbf{37.66} & \textbf{40.49} & \textbf{40.56} & 43.93 & 43.11 & 43.73 & \textbf{40.10} & \textbf{39.19} & \textbf{40.72} & 41.78 & 41.53 & 42.21 & \textbf{40.63} & 42.73 & \textbf{41.64} \\ \bottomrule 
\end{tabular}}
\end{table*}

\begin{figure*}[t!]
    \centering
    \includegraphics[width=0.95\textwidth]{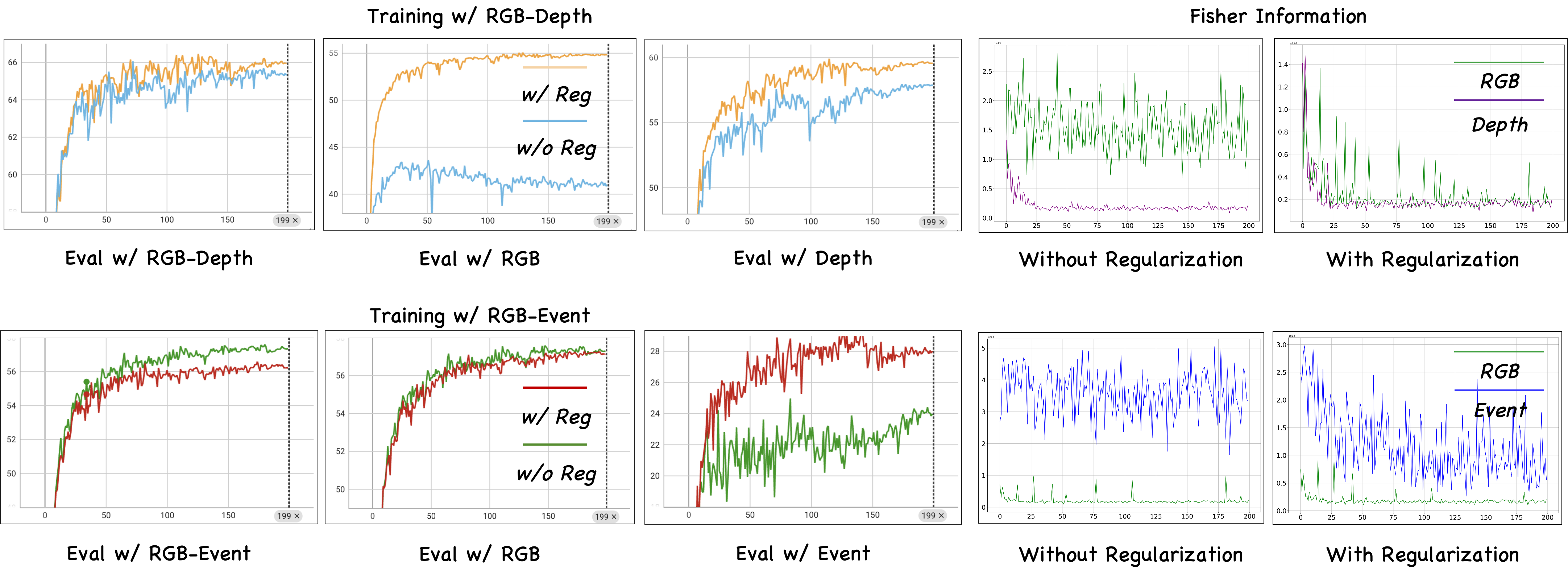}
    \caption{Comparison of model performance with (w/) and without (w/o) our proposed prediction level regularization term across different modalities (RGB-Depth, RGB, and RGB-Event).}
    \label{fig:ab}
\end{figure*}

\subsection{Results}  
Tab.~\ref{tab:delvier_RD} presents results on DELIVER~\cite{zhang2023delivering} using dual (RGB-Depth) modalities. Our method outperforms others with a mean score of 60.37\%.
Compared to MAGIC, we improve by +7.98\% in RGB-Depth fusion and +6.99\% in the overall mean, demonstrating strong performance across both modalities.  
Tab.~\ref{Tab:MUSES} shows results on MUSES~\cite{brodermann2024muses}. Our method achieves mIoU values of 60.33\% for frame, 33.15\% for event, and 42.59\% for LiDAR, with a mean mIoU of 47.80\%, surpassing prior methods. 
Compared to the previous SoTA, we achieve a mean mIoU improvement of +13.94\%, showcasing superior performance in both unimodal and multimodal tasks.  
Tab.~\ref{Tab:DELIVER} reports results on DELIVER~\cite{zhang2023delivering}. We achieve 55.56\% for RGB, 57.74\% for Depth, 32.40\% for Event, and 34.26\% for LiDAR. Paired modality combinations also perform well, with results of 59.69\% for RGB-Depth (RD), 47.70\% for RGB-Event (RE), 47.78\% for RL, 48.99\% for DE, 50.26\% for DL, and 34.37\% for EL. These results demonstrate the robustness of our method across cross-modal relationships. 
As in Tab.~\ref{Tab:MCubeS}, our method outperforms CMNeXt and MAGIC with a mIoU of 41.64\%, compared to MAGIC's 38.00\% and CMNeXt's 25.17\%. We observe significant improvements in segmentation validation for combinations such as Image and Aolp, with an average increase of +3.64\% over SoTA methods. 
Fig.~\ref{fig:visualization} shows qualitative results on DELIVER, where our method outperforms across various loss combinations and conditions. These results validate our method's ability to effectively leverage diverse modalities.

\section{Ablation}
\begin{figure}[t!]
    \centering
    \includegraphics[width=.99\linewidth]{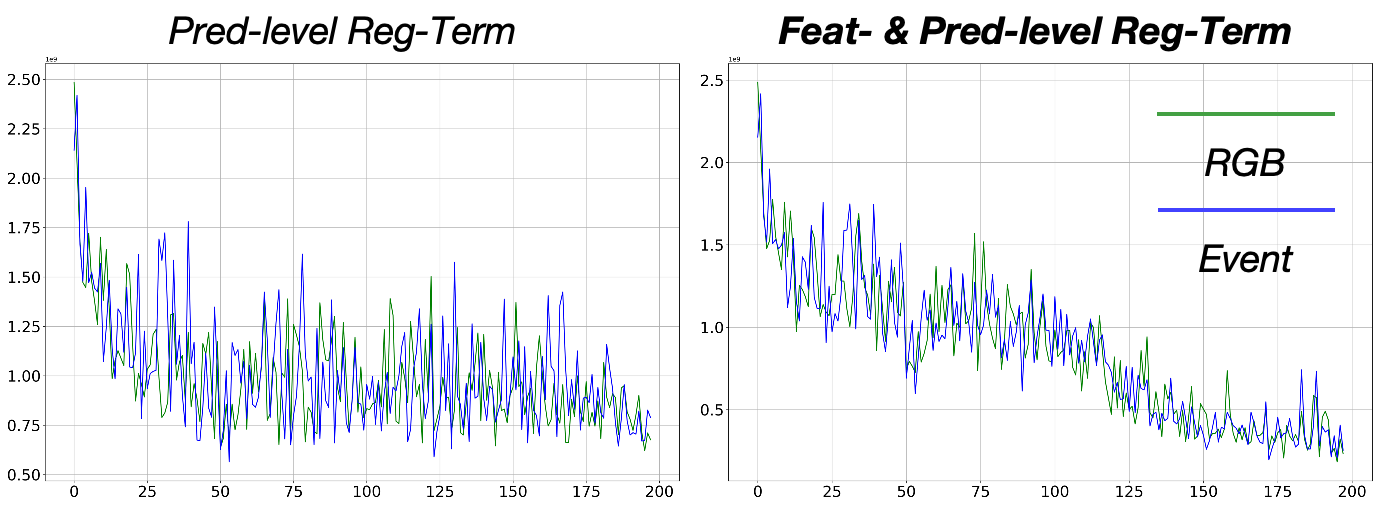}
    \caption{Ablation of the effectiveness of multi-scale feature-level regularization.
    }
    \label{fig:ab_multiscale_feat_reg}
\end{figure}

\noindent \textbf{Effectiveness of Loss Functions}  
Tab.~\ref{tab:loss_combin} presents the results of different loss function combinations on the MUSES dataset. 
Introducing \( \mathcal{L}_{Rp} \) improves performance, particularly for paired modalities. The mean mIoU increases to 50.02\%, with a +29.44\% gain for the event modality. Although the frame modality slightly decreases to 59.63\%, the overall result improves due to substantial gains in the event (+29.44\%) and both modalities (+61.00\%).  
Including \( \mathcal{L}_{Rf} \), yields a consistent performance improvement across all modalities. The mean mIoU rises to 50.44\%, with gains of +29.66\% for the frame, +61.02\% for event, and +50.44\% for both modalities. \\
\noindent \textbf{Effectiveness of Regularization Term.}  
As shown in Fig.~\ref{fig:ab}, 
the results demonstrate consistent improvement with regularization. For example, in the RGB-Depth combination, the model with regularization (w/ Reg) reaches 65\% at 150 epochs, outperforming the model without regularization (w/o Reg), which peaks at 62\%. A similar trend is observed in RGB-Event. Fisher Information plots further confirm these results, showing that without regularization, the model over-relies on individual modalities, as indicated by higher Fisher information values for RGB and Event. With regularization, Fisher information is more balanced across modalities, indicating better utilization of both inputs. This balance is particularly evident in RGB-Depth and RGB-Event combinations, where modality distances decrease with regularization. Overall, these findings highlight the effectiveness of regularization in promoting balanced modality usage, improving stability, and enhancing performance in multi-modal tasks.\\
\noindent \textbf{Ablation of Multi-scale Feature-level Regularization}
The results in Fig.~\ref{fig:ab_multiscale_feat_reg} compare the performance of two regularization strategies: the Pred-level Reg-Term (left) and the combined Feat- \& Pred-level Reg-Term (right).
The comparison clearly shows that incorporating multi-scale regularization results in a smoother reduction of Fisher information over time, as seen in the right plot. In contrast, the left plot with only prediction-level regularization exhibits more fluctuations, indicating less stable convergence.
This ablation emphasizes the significance of multi-scale feature-level regularization, as it improves stability and effectiveness, leading to smoother training dynamics and enhanced generalization across modalities.
\begin{table}[]
\caption{Ablation of different loss combinations on MUSES.}
\label{tab:loss_combin}
\renewcommand{\tabcolsep}{3pt}
\resizebox{\linewidth}{!}{
\begin{tabular}{ccc|ccc|c|ccc|c}
\midrule
\multicolumn{3}{c|}{Optimization}                                  & \multicolumn{3}{c|}{mIoU}                             & \multirow{2}{*}{Mean} & \multicolumn{3}{c|}{Acc}                             & \multirow{2}{*}{Mean} \\ \cmidrule{1-6} \cmidrule{8-10}
\multicolumn{1}{c|}{$L_{sup}$} & \multicolumn{1}{c|}{$R_p$} & $R_f$ & \multicolumn{1}{c|}{Frame} & \multicolumn{1}{c|}{Event} & Both &                       & \multicolumn{1}{c|}{Frame} & \multicolumn{1}{c|}{Event} & Both &                       \\ \midrule
\multicolumn{1}{c|}{\checkmark}           & \multicolumn{1}{c|}{-}     & -     & \multicolumn{1}{c|}{60.63}    & \multicolumn{1}{c|}{6.48} & 60.70 & 42.60                      & \multicolumn{1}{c|}{\textbf{71.34}}    & \multicolumn{1}{c|}{14.92}      & 71.32   & 52.53                      \\ \midrule
\multicolumn{1}{c|}{\checkmark}           & \multicolumn{1}{c|}{\checkmark}     & -     & \multicolumn{1}{c|}{59.63}    & \multicolumn{1}{c|}{29.44}      &  61.00  & 50.02  & \multicolumn{1}{c|}{68.89}    & \multicolumn{1}{c|}{39.39}      &  74.01  & 60.76                      \\ \midrule
\multicolumn{1}{c|}{\checkmark}           & \multicolumn{1}{c|}{\checkmark}     & \checkmark      & \multicolumn{1}{c|}{\textbf{60.63}}    & \multicolumn{1}{c|}{\textbf{29.66}}      & \textbf{61.02}   & \textbf{50.44}                      & \multicolumn{1}{c|}{69.98}    & \multicolumn{1}{c|}{\textbf{39.98}}      & \textbf{74.49}    &  \textbf{61.48}    \\ \bottomrule
\end{tabular}
}
\end{table}

\begin{table}[]
\caption{
Sensitivity analysis on the effect of different values of $\lambda_p$ on MUSES~\cite{brodermann2024muses}.}
\label{tab:ab_lambda_p}
\resizebox{\linewidth}{!}{
\begin{tabular}{c|ccc|c|ccc|c}
\midrule
\multirow{2}{*}{$\lambda_p$} & \multicolumn{3}{c|}{mIoU}                                                 & \multirow{2}{*}{Mean} & \multicolumn{3}{c|}{Acc}                                                  & \multirow{2}{*}{Mean} \\ \cmidrule{2-4} \cmidrule{6-8}
                        & \multicolumn{1}{l|}{F} & \multicolumn{1}{l|}{E} & \multicolumn{1}{l|}{FE} &                       & \multicolumn{1}{l|}{F} & \multicolumn{1}{l|}{E} & \multicolumn{1}{l|}{FE} &                       \\ \midrule
0.0 & \multicolumn{1}{c|}{57.93}  & \multicolumn{1}{c|}{28.68}  & 58.84 & 48.48 & \multicolumn{1}{c|}{67.95}  & \multicolumn{1}{c|}{37.98}  & 72.58 & 59.50                     \\ \midrule
0.1 & \multicolumn{1}{c|}{58.45}  & \multicolumn{1}{c|}{29.64}  & 59.43 & 49.17 & \multicolumn{1}{c|}{68.23}  & \multicolumn{1}{c|}{38.78}  & 73.05 & 60.02                     \\ \midrule
\rowcolor{gray!20} 0.3 & \multicolumn{1}{c|}{59.06}  & \multicolumn{1}{c|}{29.95}  & 60.67 & \textbf{49.89} & \multicolumn{1}{c|}{68.48}  & \multicolumn{1}{c|}{39.96}  & 73.93 & \textbf{60.79}                     \\ \midrule
0.5 & \multicolumn{1}{c|}{57.82}  & \multicolumn{1}{c|}{29.31}  & 59.81 & 48.98 & \multicolumn{1}{c|}{67.04}  & \multicolumn{1}{c|}{38.28}  & 73.26 & 59.53                     \\ \midrule
0.7 & \multicolumn{1}{c|}{57.60}  & \multicolumn{1}{c|}{29.04}  & 59.01 & 48.55 & \multicolumn{1}{c|}{67.19}  & \multicolumn{1}{c|}{38.89}  & 72.76 & 59.61                     \\ \bottomrule
\end{tabular}}
\end{table}

\begin{table}[]
\caption{Ablation on the effect of different $\lambda_f$ on MUSES~\cite{brodermann2024muses}.}
\label{tab:ab_lambda_f}
\resizebox{\linewidth}{!}{
\begin{tabular}{c|ccc|c|ccc|c}
\midrule
\multirow{2}{*}{$\lambda_p$} & \multicolumn{3}{c|}{mIoU}                                                 & \multirow{2}{*}{Mean} & \multicolumn{3}{c|}{Acc}                                                  & \multirow{2}{*}{Mean} \\ \cmidrule{2-4} \cmidrule{6-8}
                        & \multicolumn{1}{l|}{F} & \multicolumn{1}{l|}{E} & \multicolumn{1}{l|}{FE} &                       & \multicolumn{1}{l|}{F} & \multicolumn{1}{l|}{E} & \multicolumn{1}{l|}{FE} &                       \\ \midrule
0.0 & \multicolumn{1}{c|}{59.06}  & \multicolumn{1}{c|}{29.95}  & 60.67 & 49.89 & \multicolumn{1}{c|}{68.48}  & \multicolumn{1}{c|}{39.96}  & 73.93 & 60.79                     \\ \midrule
\rowcolor{gray!20} 0.02 & \multicolumn{1}{c|}{60.63}  & \multicolumn{1}{c|}{29.66}  & 61.02 & \textbf{50.44} & \multicolumn{1}{c|}{69.88}  & \multicolumn{1}{c|}{39.98}  & 74.49 & \textbf{61.45}                     \\ \midrule
0.04 & \multicolumn{1}{c|}{59.96}  & \multicolumn{1}{c|}{30.09}  & 60.74 & 50.26 & \multicolumn{1}{c|}{69.38}  & \multicolumn{1}{c|}{40.44}  & 74.29 & 61.37                     \\ \midrule
0.06 & \multicolumn{1}{c|}{58.84}  & \multicolumn{1}{c|}{31.08}  & 60.16 & 50.03 & \multicolumn{1}{c|}{68.62}  & \multicolumn{1}{c|}{40.96}  & 74.05 & 61.21                     \\ \bottomrule
\end{tabular}}
\end{table}

\noindent \textbf{Hyper-parameters}
Tables~\ref{tab:ab_lambda_p} and~\ref{tab:ab_lambda_f} show the impact of $\lambda_p$ and $\lambda_f$ on the MUSES dataset. 
In Table~\ref{tab:ab_lambda_p}, $\lambda_p = 0.3$ achieves the best performance with a mIoU of 59.06\% and accuracy of 60.79\%, outperforming other values. Lower values (0.0 and 0.1) result in slightly worse performance, while higher values (0.5 and 0.7) show a decrease, especially in frame modality.
Table~\ref{tab:ab_lambda_f} demonstrates that $\lambda_f = 0.02$ gives the best results, with a mIoU of 60.63\% and accuracy of 61.45\%. Larger $\lambda_f$ values (0.04 and 0.06) cause a slight performance drop.
These findings highlight the importance of tuning $\lambda_p$ and $\lambda_f$, with $\lambda_p = 0.3$ and $\lambda_f = 0.02$ yielding the best results.

\section{Discussion}
\noindent \textbf{Differences with ~\cite{gat2020removing}.} Unlike ~\cite{gat2020removing}, which focuses on image-level visual question answering with image and text modalities, we address the more complex pixel-level task by incorporating multiple visual modalities. Challenges arise from both improving modality utilization and balancing multiple modalities. When we directly apply the regularization term from ~\cite{gat2020removing}, the resulting mIoU on MUSES is 42.89\%, significantly lower than our 50.44\%.

\begin{figure}[t!]
    \centering
    \includegraphics[width=.99\linewidth]{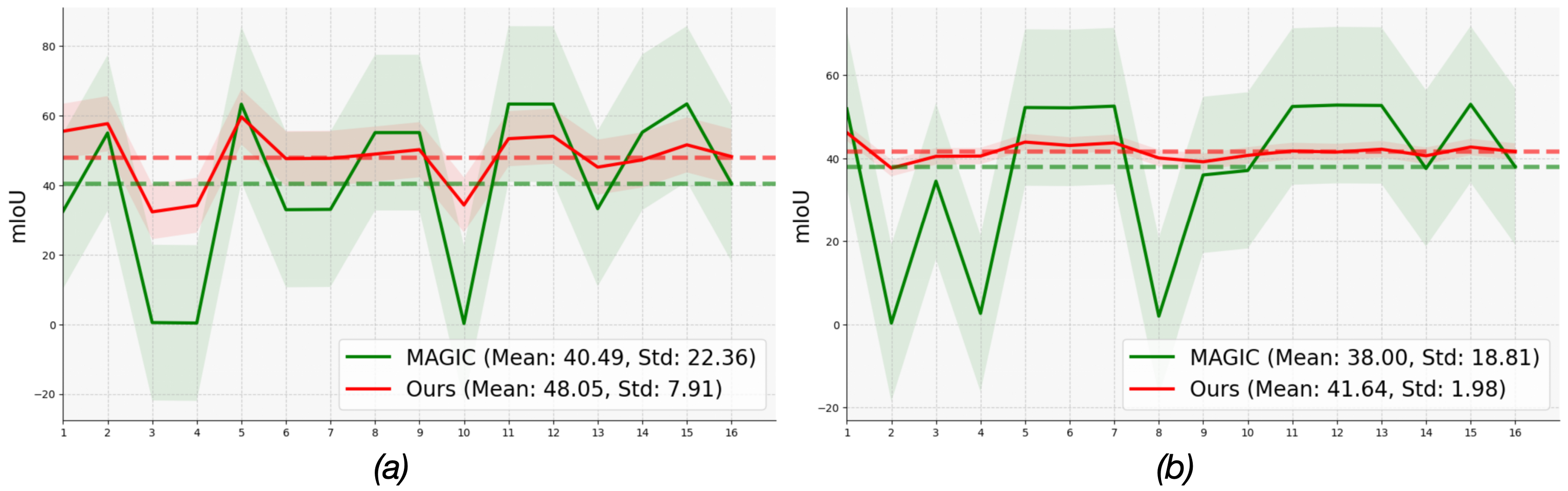}
    \caption{Balanced multi-modal performance comparison between ours and MAGIC~\cite{zheng2025centering} on (a) DELIVER and (b) MCubeS.
    }
    \label{fig:balanced_ab}
\end{figure}

\noindent \textbf{Unimodal Bias.} As in Fig.~\ref{fig:balanced_ab}, compared to MAGIC~\cite{zheng2025centering} (green line), which exhibits a notable unimodal bias, our method (red line) demonstrates consistent performance across modality combinations. MAGIC shows a mean mIoU of 40.49\% with a high standard deviation of 22.36, reflecting sensitivity to certain modality pairs. In contrast, our method achieves a higher mean mIoU of 48.05\% with a smaller standard deviation of 7.91, highlighting its robustness and stability. 
This trend emphasizes the absence of unimodal bias in our approach, as it handles all modality combinations effectively, unlike MAGIC, which suffers from significant fluctuations. The shaded areas around the curves further emphasize the reduced variability of our method, demonstrating its stability and reliability in achieving balanced performance across modalities.

\begin{table}[]
\caption{Ablation of training efficiency with 8 H100 GPUs.}
\label{tab:ab_training_ef}
\renewcommand{\tabcolsep}{14pt}
\resizebox{\linewidth}{!}{
\begin{tabular}{c|c|c}
\toprule
              & w/o Reg Terms & w/ Reg Terms \\ \midrule
Training Time (hours) &  2.462             &  2.476  (0.014$\uparrow$)          \\ 
\bottomrule
\end{tabular}}
\end{table}

\noindent \textbf{Training Efficiency}  
As shown in Tab.~\ref{tab:ab_training_ef}, we compare the training efficiency of the model with and without regularization terms across 8 NVIDIA GPUs. The results show a slight increase in training time, with the model using regularization taking 2.476 hours, compared to 2.462 hours without regularization—an increase of 0.014 hours (approximately 0.57\%). Despite this minimal increase, the performance improvements justify the inclusion of regularization, offering a favorable trade-off between efficiency and accuracy. Thus, regularization proves to be a valuable investment with negligible impact on training time.

\section{Conclusion}
We presented a simple yet effective regularization term based on functional entropy to mitigate unimodal bias in multi-modal semantic segmentation. Our approach ensures balanced utilization of all input modalities, improving robustness in scenarios where some modalities are unavailable or degraded. Extensive experiments on synthetic and real-world benchmarks demonstrated significant performance improvements, achieving better results without introducing extra parameters or complexity. 
Key contributions include introducing a novel regularization technique, proposing a simple fusion framework, and applying entropy maximization for balanced multi-modal learning. 
Our method outperforms existing approaches, offering a practical solution for robust multi-modal integration with multiple visual modalities. Future work could extend this approach to other multi-modal tasks for robust and stable scene understanding.
{
    \small
    \bibliographystyle{ieeetr}
    \bibliography{main}
}

\end{document}